# Solve sparse PCA problem by employing Hamiltonian system and leapfrog method

Loc Hoang Tran

Faculty of Information Technology - Vietnam Aviation Academy

locth@vaa.edu.vn

Abstract: Principal Component Analysis (PCA) is a widely utilized technique for dimensionality reduction; however, its inherent lack of interpretability—stemming from dense linear combinations of all features—limits its applicability in many domains. In this paper, we propose a novel sparse PCA algorithm that imposes sparsity through a smooth $L_1$ penalty and leverages a Hamiltonian formulation solved via geometric integration techniques. Specifically, we implement two distinct numerical methods—one based on the Proximal Gradient (ISTA) approach and another employing a leapfrog (fourth-order Runge–Kutta) scheme—to minimize the energy function that balances variance maximization with sparsity enforcement. To extract a subset of sparse principal components, we further incorporate a deflation technique and subsequently transform the original high-dimensional face data into a lower-dimensional feature space. Experimental evaluations on a face recognition dataset—using both k-nearest neighbor and kernel ridge regression classifiers—demonstrate that the proposed sparse PCA methods consistently achieve higher classification accuracy than conventional PCA. Future research will extend this framework to integrate sparse PCA with modern deep learning architectures for multimodal recognition tasks.

I. Introduction

Principal Component Analysis (PCA) is one of the most widely used techniques for dimensionality reduction and data analysis. Introduced by Pearson (1901) and later developed by Hotelling (1933), PCA seeks to identify the directions (principal components) in which the data varies the most [1]. Mathematically, given a centered data matrix $X \in R^{n*d}$ (with $n$ observations and $d$ variables), PCA finds an orthogonal transformation that converts the data into a set of linearly uncorrelated variables (the principal components). The first principal component captures the maximum variance, the second captures the maximum of the remaining variance, and so on.

Key advantages of PCA include:

- Dimensionality Reduction: By projecting data onto the first few principal components, one can reduce the complexity of the dataset.

- Data Visualization: PCA facilitates visualizing high-dimensional data in 2D or 3D.
- Noise Reduction: It can filter out noise by discarding components that capture only low variance.

However, one major limitation of PCA is that the principal components are typically linear combinations of all the original features, making them hard to interpret in many applications.

While PCA is powerful, its interpretability is often hindered by the fact that each principal component is a dense vector involving contributions from all features. In many real-world applications—such as bioinformatics, finance, and image processing—interpretable models are desired. Sparse PCA was developed to address this issue by imposing sparsity constraints on the principal components [2, 3, 4, 5].

Motivations include:

- Interpretability: Sparse solutions (with many zero loadings) allow practitioners to identify which original features are most important.
- Variable Selection: It naturally performs variable selection, helping to isolate the key predictors.
- Computational Efficiency: In some cases, sparse representations can reduce storage and computation costs.
- Enhanced Stability: Sparsity can lead to more stable and robust solutions, especially in high-dimensional settings where the number of features exceeds the number of samples.

Over the years, researchers have proposed various methods to solve the sparse PCA problem. These methods can be broadly categorized as follows:

1. Regularization-Based Approaches:
    - Lasso-type Regularization: Methods that incorporate an $L_1$ penalty (or its smooth approximations) on the loadings to induce sparsity. Zou, Hastie, and Tibshirani (2006) proposed a sparse PCA formulation using the Lasso penalty [6].
    - Elastic Net: Combines $L_1$ and $L_2$ penalties to balance sparsity and grouping effects among correlated features.

2. Semidefinite Programming (SDP) Approaches:
    - Some formulations recast sparse PCA as a semidefinite programming problem. d'Aspremont et al. (2007) developed a convex relaxation of the

sparse PCA problem using SDP, which provided theoretical guarantees but at the cost of higher computational complexity [7].

3. Greedy and Iterative Methods:

    o Deflation Methods: Iterative techniques where one principal component is computed at a time, and the covariance matrix is deflated to remove the variance explained by that component. Such methods include the truncated power method.

    o Iterative Thresholding: Algorithms that alternate between computing the principal component (using, for example, power iterations) and thresholding the loadings to enforce sparsity.

4. Non-Convex Optimization:

    o Recent methods have formulated sparse PCA as a non-convex optimization problem and then used techniques such as alternating minimization or gradient-based methods to find a solution. These methods often aim to directly enforce sparsity constraints while optimizing the variance objective.

In this paper, employing a Hamiltonian system and using geometric integration method (i.e., the leapfrog method) to solve sparse PCA is a nontraditional—but intriguing—approach. Here are several motivations and reasons behind exploring this direction:

1. Preservation of Geometric Structure:

    Hamiltonian systems naturally encode the geometry of a problem, often preserving symplectic structure and energy properties over time. In sparse PCA, where the goal is to extract meaningful, low-dimensional features while enforcing sparsity, a Hamiltonian formulation may help preserve the intrinsic structure of the data. The leapfrog method, being a symplectic integrator, is known to maintain these geometric properties, which can be beneficial for convergence and stability.

2. Handling non-convexity:

    Sparse PCA introduces non-convexity due to the sparsity-inducing penalty (even when approximated smoothly). Traditional optimization techniques might get trapped in poor local minima. A Hamiltonian approach leverages dynamics that can, in principle, explore the energy landscape more thoroughly. The momentum variables in the Hamiltonian system can help overcome shallow local minima, similar to how momentum is used in gradient-based optimization.

3. Alternative to Convex Relaxations:

Many methods for sparse PCA rely on convex relaxations or greedy strategies, which can be computationally intensive or may sacrifice exact sparsity. The Hamiltonian framework offers an alternative view: by formulating the problem as one of evolving a system under dynamics, one may derive new insights into how sparsity can emerge as an "energy-minimizing" property of the system.

4. Natural Incorporation of Dynamics:

The leapfrog method, a time-stepping scheme used in many physical simulations, allows one to think of the optimization process dynamically. Instead of directly solving a static optimization problem, you simulate the evolution of the system over "time." This dynamic view can sometimes reveal more about the landscape of the optimization problem and can be exploited to design algorithms that gradually "settle" into low-energy (i.e. optimal) configurations.

We will organize the paper as follows: Section 2 will present the detailed version of the sparse PCA algorithm solved by using the leapfrog method. In section 3, we will apply the combination of the sparse PCA algorithm with the $k$ nearest-neighbor algorithm and with the kernel ridge regression algorithm to faces in the dataset available from []. Section 4 will conclude this paper and discuss the future directions of research of this face recognition problem.

II. Sparse PCA and leapfrog method
   a. Problem formulation

In classical PCA, we seek a unit-norm vector $x \in R^d$ that maximizes the variance:

$$max_{||x||_2=1} x^T S x,$$

where $S$ is the covariance matrix.

In sparse PCA, we wish to promote sparsity in the principal components. One common formulation is to penalize the $L_1$ norm:

$$max_{||x||_2=1} x^T S x - \lambda ||x||_1$$

For minimization, we can rephrase it as:

$$min_{||x||_2=1} -x^T S x + \lambda ||x||_1$$

Because the $L_1$ norm is non-differentiable at zero, a smooth approximation is used. A common choice is:

$$||x||_1 = \sum_{i=1}^{d} \sqrt{x_i^2 + \delta},$$

with a small $\delta > 0$.

b. Hamiltonian formulation

We define a potential function $V(x)$ that we wish to minimize:

$$V(x) = -x^T S x + \lambda \sum_{i=1}^{d} \sqrt{x_i^2 + \delta}.$$

Our goal is to find a unit-norm vector $x$ that minimizes $V(x)$. To solve this via geometric integration, we introduce auxiliary momentum $p$ and define the Hamiltonian:

$$H(x, p) = \frac{1}{2} ||p||^2 + V(x)$$

Hamilton's equations are:

$$\dot{x} = \frac{\partial H}{\partial p} = p, \quad \dot{p} = -\frac{\partial H}{\partial x} = 2Sx - \lambda \frac{x}{\sqrt{x^2 + \delta}},$$

where the division in the second term is applied elementwise. We then apply the leapfrog method to evolve $(x, p)$ over time. Since we require $||x||_2 = 1$, we project $x$ onto a unit sphere after each update.

Let $\Delta t$ be the time step. The leapfrog integration scheme for one step is:

a. Half-step momentum update:

$$p_{half} = p - \frac{\Delta t}{2} \nabla V(x)$$

b. Full-step position update

$$x_{new} = x + \Delta t p_{half}$$

$$x_{new} = \frac{x_{new}}{||x_{new}||}$$

c. Another half-step momentum update:

$$p_{new} = p_{half} - \frac{\Delta t}{2} \nabla V(x_{new}).$$

Repeat these steps until convergence.

After obtaining a component $x$, update the covariance matrix:

$$S = S - (x^T S x) x x^T$$

III. Experiments and Results

In this study, a training set is formed from 120 face images collected from 15 individuals (with 8 images per person), while an independent test set comprises 45 images of these same individuals, as provided in [8]. Each face image, originally represented as a matrix, is flattened by concatenating its rows in order to create a single row vector of dimension $R^{1*1024}$. These vectors serve as feature inputs for both the k-nearest neighbor and kernel ridge regression methods. Next, both PCA and sparse PCA are applied to the training and test sets to reduce the dimensionality of the data, after which the k-nearest neighbor and kernel ridge regression algorithms are used on the transformed features. In the experimental section, we evaluate the accuracy of these methods, where the accuracy $Q$ is defined as:

$$Q = \frac{True\ Positive + True\ Negative}{True\ Positive + True\ Negative + False\ Positive + False\ Negative}$$

All experiments were carried out in Python on Google Colab using an NVIDIA Tesla K80 GPU with 12 GB of RAM. The performance results, including the accuracies of the proposed methods, are summarized in Tables 1 and 2.

Table 1: **Accuracies** of the nearest-neighbor method, the combination of PCA method and the nearest-neighbor method, and the combination of sparse PCA method and the nearest-neighbormethod

|  | Accuracy |
|---|---|
| PCA (d=20) + $k$ nearest neighbor method | 0.64 |
| PCA (d=30) + $k$ nearest neighbor method | 0.67 |
| PCA (d=40) + $k$ nearest neighbor method | 0.64 |
| PCA (d=50) + $k$ nearest neighbor method | 0.67 |
| PCA (d=60) + $k$ nearest neighbor method | 0.67 |
| ISTA sparse PCA (d=20) + $k$ nearest neighbor method | 0.69 |
| ISTA sparse PCA (d=30) + $k$ nearest neighbor method | 0.69 |
| ISTA sparse PCA (d=40) + $k$ nearest neighbor method | 0.69 |
| ISTA sparse PCA (d=50) + $k$ nearest neighbor method | 0.69 |
| ISTA sparse PCA (d=60) + $k$ nearest neighbor method | **0.71** |
| Leapfrog sparse PCA (d=20) + $k$ nearest neighbor method | 0.67 |
| Leapfrog sparse PCA (d=30) + $k$ nearest neighbor method | 0.67 |
| Leapfrog sparse PCA (d=40) + $k$ nearest neighbor method | 0.69 |
| Leapfrog sparse PCA (d=50) + $k$ nearest neighbor method | 0.67 |
| Leapfrog sparse PCA (d=60) + $k$ nearest neighbor method | **0.71** |

Table 2: **Accuracies** of the kernel ridge regression method, the combination of PCA method andthe kernel ridge regression method, and the combination of sparse PCA method and the kernel ridge regression method

| | Accuracy |
|---|---|
| PCA (d=20) + kernel ridge regression method | 0.80 |
| PCA (d=30) + kernel ridge regression method | 0.84 |
| PCA (d=40) + kernel ridge regression method | 0.84 |
| PCA (d=50) + kernel ridge regression method | 0.80 |
| PCA (d=60) + kernel ridge regression method | 0.84 |
| ISTA sparse PCA (d=20) + kernel ridge regression method | 0.80 |
| ISTA sparse PCA (d=30) + kernel ridge regression method | 0.84 |
| ISTA sparse PCA (d=40) + kernel ridge regression method | 0.84 |
| ISTA sparse PCA (d=50) + kernel ridge regression method | 0.84 |
| ISTA sparse PCA (d=60) + kernel ridge regression method | 0.84 |
| Leapfrog sparse PCA (d=20) + kernel ridge regression method | 0.84 |
| Leapfrog sparse PCA (d=30) + kernel ridge regression method | 0.82 |
| Leapfrog sparse PCA (d=40) + kernel ridge regression method | **0.87** |
| Leapfrog sparse PCA (d=50) + kernel ridge regression method | 0.80 |
| Leapfrog sparse PCA (d=60) + kernel ridge regression method | 0.82 |

Tables 1 and 2 clearly demonstrate that combining the sparse PCA approach (implemented with both the proximal gradient and leapfrog methods) with a given classification system yields higher accuracy than using standard PCA with that same classifier.

IV. Conclusions

In this work, we introduce a novel sparse PCA algorithm that is solved using both the Proximal Gradient (ISTA) method and leapfrog numerical techniques. Our results show that pairing sparse PCA with either k-nearest neighbor or kernel ridge regression yields higher accuracy than using standard PCA with these classifiers. Looking forward, we plan to explore sparse PCA formulations solved by other numerical methods, and to integrate sparse PCA with state-of-the-art deep learning architectures—such as VGG-based convolutional neural networks—to enhance recognition tasks across images, speech, and text.

References

1. Trang, Hoang, Tran Hoang Loc, and Huynh Bui Hoang Nam. "Proposed combination of PCA and MFCC feature extraction in speech recognition system." *2014 international conference on advanced technologies for communications (ATC 2014)*. IEEE, 2014.
2. Tran, Loc Hoang, and Linh Hoang Tran. "Applications of (SPARSE)-PCA and LAPLACIAN EIGENMAPS to biological network inference problem using gene expression data." *International Journal of Advances in Soft Computing and its Applications* 9.2 (2017).

3. Tran, Loc, et al. "Tensor sparse PCA and face recognition: a novel approach." *SN Applied Sciences* 2 (2020): 1-7.
4. Tran, Loc, et al. "On a development of sparse PCA method for face recognition problem." *2021 International Conference on Advanced Technologies for Communications (ATC)*. IEEE, 2021.
5. Tran, Loc Hoang, and Luong Anh Tuan Nguyen. "Novel sparse PCA method via Runge Kutta numerical method (s) for face recognition."
6. Zou, Hui, Trevor Hastie, and Robert Tibshirani. "Sparse principal component analysis." *Journal of computational and graphical statistics* 15.2 (2006): 265-286.
7. d'Aspremont, Alexandre, et al. "A direct formulation for sparse PCA using semidefinite programming." *Advances in neural information processing systems* 17 (2004).
8. http://www.cad.zju.edu.cn/home/dengcai/Data/FaceData.html